\pdfoutput=1

\documentclass[11pt]{article}
\usepackage{longtable} 
\usepackage{array}     
\usepackage{booktabs}  

\usepackage[final]{acl}

\usepackage{times}
\usepackage{latexsym}

\usepackage[T1]{fontenc}

\usepackage[utf8]{inputenc}

\usepackage{microtype}

\usepackage{inconsolata}

\usepackage{graphicx}
\usepackage{array} 

\usepackage{amsmath}  
\usepackage{amsfonts} 

\usepackage{latexsym}
\usepackage{booktabs} 

\usepackage{tabularx}
\title{Transparent but Powerful: Explainability, Accuracy, and Generalizability in ADHD Detection from Social Media Data}

\author{
    Daniel Wiechmann$^{1,3}$, Edward Kempa$^{2}$, Elma Kerz$^{3}$, Yu Qiao$^{3}$ \\
    $^{1}$Institute for Logic, Language \& Computation, University of Amsterdam, the Netherlands\\
    $^{2}$Department of Computer and Information Science and Engineering, University of Florida, USA\\
    $^{3}$Exaia Technologies GmbH, Germany\\
    \texttt{\{d.wiechmann,e.kerz,y.qiao\}@exaia-tech.com},\\ \texttt{d.wiechmann@uva.nl}, \texttt{kempaedward@ufl.edu} \\
}

\begin{document}
\maketitle
\begin{abstract}
Attention-deficit/hyperactivity disorder (ADHD) is a prevalent mental health condition affecting both children and adults, yet it remains severely underdiagnosed. Recent advances in artificial intelligence, particularly in Natural Language Processing (NLP) and Machine Learning (ML), offer promising solutions for scalable and non-invasive ADHD screening methods using social media data. This paper presents a comprehensive study on ADHD detection, leveraging both shallow machine learning models and deep learning approaches, including BiLSTM and transformer-based models, to analyze linguistic patterns in ADHD-related social media text. Our results highlight the trade-offs between interpretability and performance across different models, with BiLSTM offering a balance of transparency and accuracy. Additionally, we assess the generalizability of these models using cross-platform data from Reddit and Twitter, uncovering key linguistic features associated with ADHD that could contribute to more effective digital screening tools.

\end{abstract}

\section{Introduction}

Attention-deficit/hyperactivity disorder (ADHD) is one of the most prevalent mental health disorders, impacting both children and adults. It is characterized by persistent patterns of inattention, impulsivity, and hyperactivity \cite[DSM-5-TR,][]{APA2022}. In children, the estimated prevalence ranges from 5\% to 13\%, depending on the diagnostic screening methods employed. In adults, the prevalence is lower yet significant, with estimates varying from 2.5\% to 4.4\% of the population \cite{polanczyk2007worldwide,ayano2023prevalence}. A study by \citet{chung2019trends} reported that the prevalence of adult ADHD in the United States had risen to 0.96\%, doubling from 0.43\% a decade earlier. More recently, a national parental survey indicated that approximately 7 million children aged 3–17 years in the United States (11.4\%) have been diagnosed with ADHD \cite{danielson2024adhd}. The cognitive impairments associated with ADHD can have long-lasting effects \cite{kendall2008diagnosis}, influencing academic performance, career success, and overall quality of life. Despite the significant personal and societal burden associated with ADHD, similar to other psychiatric conditions, ADHD is severely underdiagnosed.  Less than half of the persons with ADHD symptoms are believed to have ever received a clinical diagnosis \cite{chamberlain2017adhd}.

Recent advances in artificial intelligence have spurred research into cost-effective, non-intrusive, and scalable screening procedures for mental health disorders. In particular, the combination of natural language processing (NLP) and machine learning (ML) harnessing textual data from social media is increasingly recognized for its transformative potential in supporting healthcare professionals in the early detection, treatment, and prevention of mental disorders, thus empowering proactive mental healthcare \cite[see][, for comprehensive reviews]{calvo2017natural,Zhang2022,zhou2022natural}. Social media offers a unique opportunity for the collection of naturalistic and ecologically valid data on the daily experiences and public expressions of individuals with ADHD. By analyzing digital footprints of verbal behavior on social media platforms, researchers can extract meaningful information about individuals’ feelings and symptoms. 

The existing literature has extensively addressed conditions such as depression, suicidal ideation, and anxiety; however, research on attention-deficit/hyperactivity disorder (ADHD) remains comparatively limited \cite[for exceptions, see][]{guntuku2019language}. Current feature-engineered approaches often adopt an opportunistic selection of features to identify potential digital biomarkers for ADHD, predominantly utilizing dictionary-based methods like the Linguistic Inquiry and Word Count (LIWC) dictionary. Conversely, the most predictive approaches relying on transformer-based architectures are often black-box in nature, obscuring the underlying mechanisms of the model and hindering transparency \cite[see][for an exception]{kerz2023toward}. This lack of transparency is particularly problematic for the adaptation of ADHD machine learning models in high-stakes and sensitive real-world applications, where understanding the decision-making process is crucial.

Moreover, there is a notable absence of efforts to assess the generalizability of ADHD detection models across diverse populations and contexts. Research by \citet{harrigian-etal-2020-models} highlights the limitations in the generalizability of current models, particularly concerning depression detection. This underscores the need for a more robust framework that enhances the accuracy of ADHD detection and ensures reliable applicability across varied settings.

To address these challenges, our work makes several key contributions:

\begin{itemize}
    \item \textbf{Model Development}: We developed a suite of ADHD detection models, including both shallow machine learning methods and a BiLSTM, trained on a comprehensive set of human-interpretable features that capture ADHD-related linguistic patterns across eight dimensions of verbal behavior.
    \item \textbf{Interpretability vs. Performance}: We evaluate the trade-off between interpretability and performance by comparing our interpretable models to both domain-general and domain-adapted fine-tuned transformer models.
    \item \textbf{Feature Ablation}: We conduct feature ablation experiments to identify the most informative feature groups that are strongly associated with ADHD.
    \item \textbf{Generalizability Assessment}: We assess model generalizability through out-of-distribution experiments, training on Reddit data and testing on Twitter data.

\end{itemize}

\section{Related Work}

Due to the significant shortcomings and limitations of current ADHD screening procedures, along with the absence of large-scale screening methods, emerging approaches and technologies are attracting increasing attention within the scientific community. Among these, cognitive Event-Related Potentials (ERPs) within electroencephalograms (EEG) have demonstrated robust neurophysiological distinctions between individuals diagnosed with ADHD and those without. Studies have reported differences in brain structural and functional measures related to cognitive functions in ADHD patients \cite{hoogman2019brain, lake2019functional}. However, these methods are constrained by high costs and the requirement for specialized technical expertise \cite{van2010error}. Another emerging technology, gaze eye-tracking technology, also shows potential for ADHD detection \cite{fried2014adhd,lee2023use}. However, its application is limited by the need for specialized equipment such as high-speed cameras, which are typically confined to research labs, making widespread clinical implementation challenging. The collection of high-quality eye-tracking data is resource-intensive and often results in a scarcity of large datasets necessary for robust research and model development. These technical and methodological challenges emphasize the need for more accessible and reliable integrated diagnostic approaches for ADHD screening.

Another emerging approach in the field leverages Natural Language Processing (NLP) and Machine Learning (ML) to analyze social media text for ADHD detection, commonly referred to as Mental Illness Detection and Analysis on Social Media (MIDAS). MIDAS is an interdisciplinary field situated at the intersection of multiple disciplines, including Computational Linguistics, Computational Social Science, Cognitive Psychology, and Clinical Psychiatry. Within this field, automated detection of mental health conditions is typically approached as a classification task or sentiment analysis, where NLP techniques are used to extract linguistic, statistical, and domain features from social media data. These features are then fed into supervised machine learning models to predict the presence of specific mental disorders and symptomatology. This approach holds significant potential for the development of a digital phenotype, a computationally derived characterization of an individual that can be analyzed for signs of mental illness, allowing for early detection and intervention \cite{liang2019survey, zhang2022natural, garg2023mental}.

\section{Experimental Setup}

In this section, we first introduce the datasets, then (1) provide the implementation details for all models, including the transformer-based baselines, and analyze their performance in a in-domain setting (training and testing on Reddit data), (2) assess the generalizability of the models by evaluating their performance in an out-of-domain setting (training on Reddit data, testing on Twitter data), and (3) explore the linguistic manifestations of ADHD through feature ablation experiments, identifying the most informative features and feature groups contributing to the detection of the disorder.

\subsection{Dataset}

We constructed an ADHD dataset by reimplementing and refining the data collection method of SMHD \cite{cohan-etal-2018-smhd}. Users and posts were extracted from a publicly available Reddit corpus using the Arctic Shift Project\footnote{\url{https://github.com/ArthurHeitmann/arctic_shift}}. Diagnosed ADHD users were identified based on high-precision detection patterns. These patterns consist of two components: one that matches self-reported diagnosis phrases (e.g., “diagnosed with”), and another that maps relevant ADHD-related keywords (e.g., “attention deficit,” “hyperactivity”). A user was labeled with ADHD if one of these keywords appears within 40 characters of the diagnosis pattern. Control users were randomly sampled from those who never posted or commented in mental health-related subreddits and never mentioned ADHD or similar mental health terms, in order to minimize false positives. Following SMHD's methodology, we removed posts that directly referenced ADHD diagnoses to prevent label leakage, while retaining mental health-related posts to allow for accurate symptom feature extraction. Pre-processing procedures applied to the final dataset are detailed in the Appendix. The final dataset consists of 12k diagnosed ADHD users and a matched number of control users. The full dataset statistics are presented in Table \ref{tab:adhd_stats}.

\begin{table}[h!]
    \centering
    \label{tab:adhd_stats}
    \begin{tabularx}{\columnwidth}{Xrrr}  
        \toprule
        User group & N users & N texts & N sentences \\
        \midrule
        ADHD       & 12,070   & 317,073  & 2,830,661 \\
        Control    & 12,070   & 174,765  & 1,266,155 \\
        \bottomrule
    \end{tabularx}
    \caption{ADHD Dataset Statistics}
\label{tab:adhd_stats}
\end{table}


\subsection{A feature framework for ADHD detection models}

Feature extraction was conducted using EXAIA CYMO V1.0.0-beta, a proprietary text analytics and mining tool designed to integrate a comprehensive set of expert-engineered features that capture multiple dimensions of language use. The current version of CYMO supports 344 features categorized into eight distinct groups: (1) Syntactic Complexity, (2) Lexical Richness/Complexity, (3) Cohesion, (4) Stylistics, (5) Readability, (6) Grammatical Categories, (7) Topical Categories, and (8) Emotion Categories. The \textbf{Syntactic Complexity} group encompasses five types of measures: (a) length of production unit, (b) sentence complexity, (c) subordination, (d) coordination, and (e) specific structures. The \textbf{Lexical Richness/Complexity} group addresses the diversity of vocabulary, with measures including (a) lexical diversity, (b) lexical sophistication, (c) lexical density, and (d) word prevalence. \textbf{Cohesion} pertains to cues that connect ideas within a text, specifically focusing on (a) lexical overlap and (b) the use of connectives. The \textbf{Stylistics} group examines variations in language use by register, genre, and style, operationalized through register/genre-specific n-gram measures that consider frequency and count across multiple contexts. \textbf{Readability} measures, such as the Flesch-Kincaid Grade Level, evaluate the ease of understanding a text by assessing factors like sentence length and grammatical complexity. The \textbf{Grammatical Categories} group is dictionary-based, encompassing prepositions, determiners, auxiliary verbs, pronouns, conjunctions, and quantifiers. \textbf{Topical Categories} cover diverse domains including Art, Business, Education, Entertainment, Fashion, Food, Health, Music, Politics, Relationships, Science, Sports, Technology, and Travel. Lastly, the \textbf{Emotion Categories} incorporate both positive and negative emotions based on established psychological models, as detailed in the revised Hourglass Model by \citet{susanto2020hourglass}.\footnote{A detailed feature list can be found at this URL \url{https://cymo-doc.exaia-tech.net/}.} 

\begin{figure*}[t]
  \includegraphics[width=0.48\linewidth]{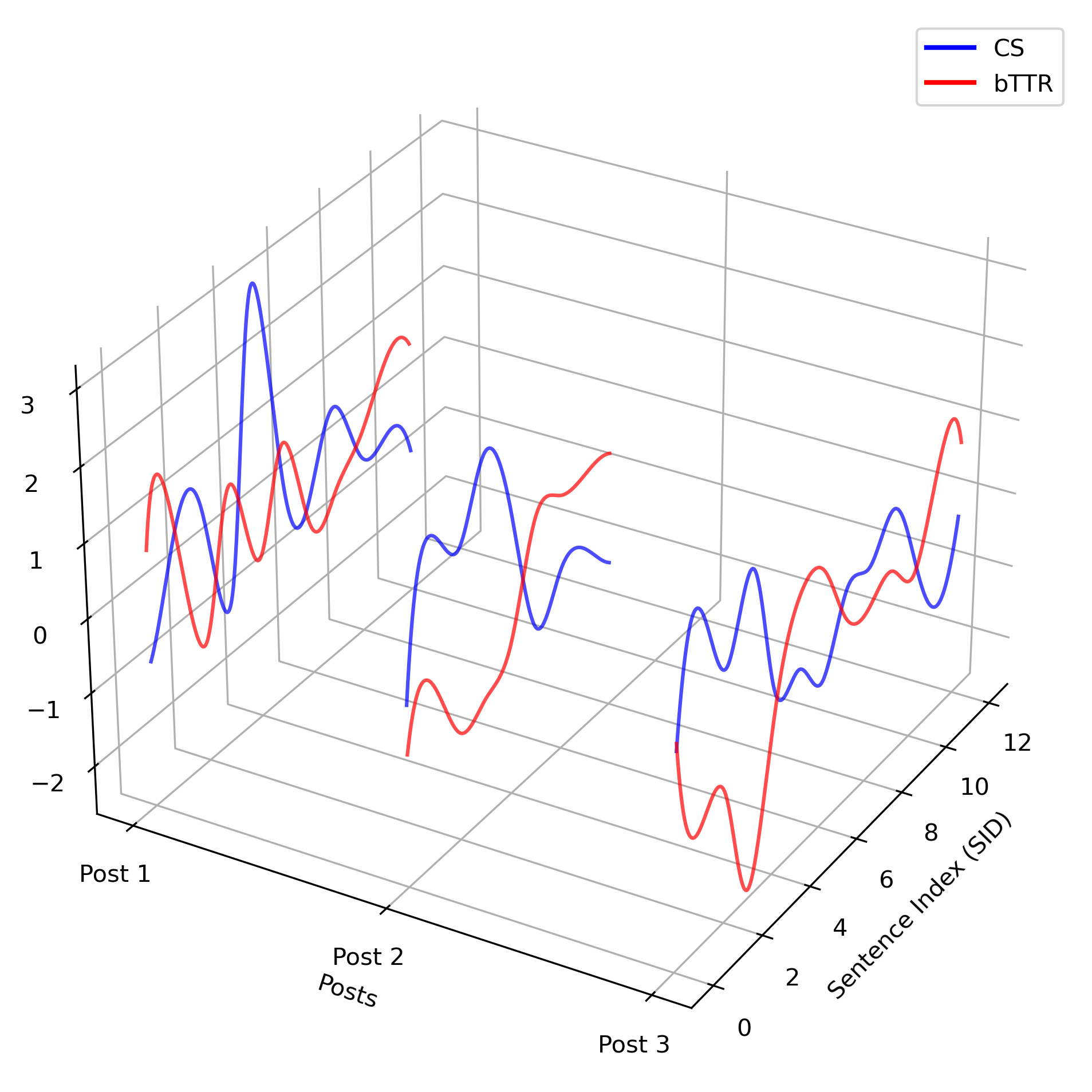} \hfill
  \includegraphics[width=0.48\linewidth]{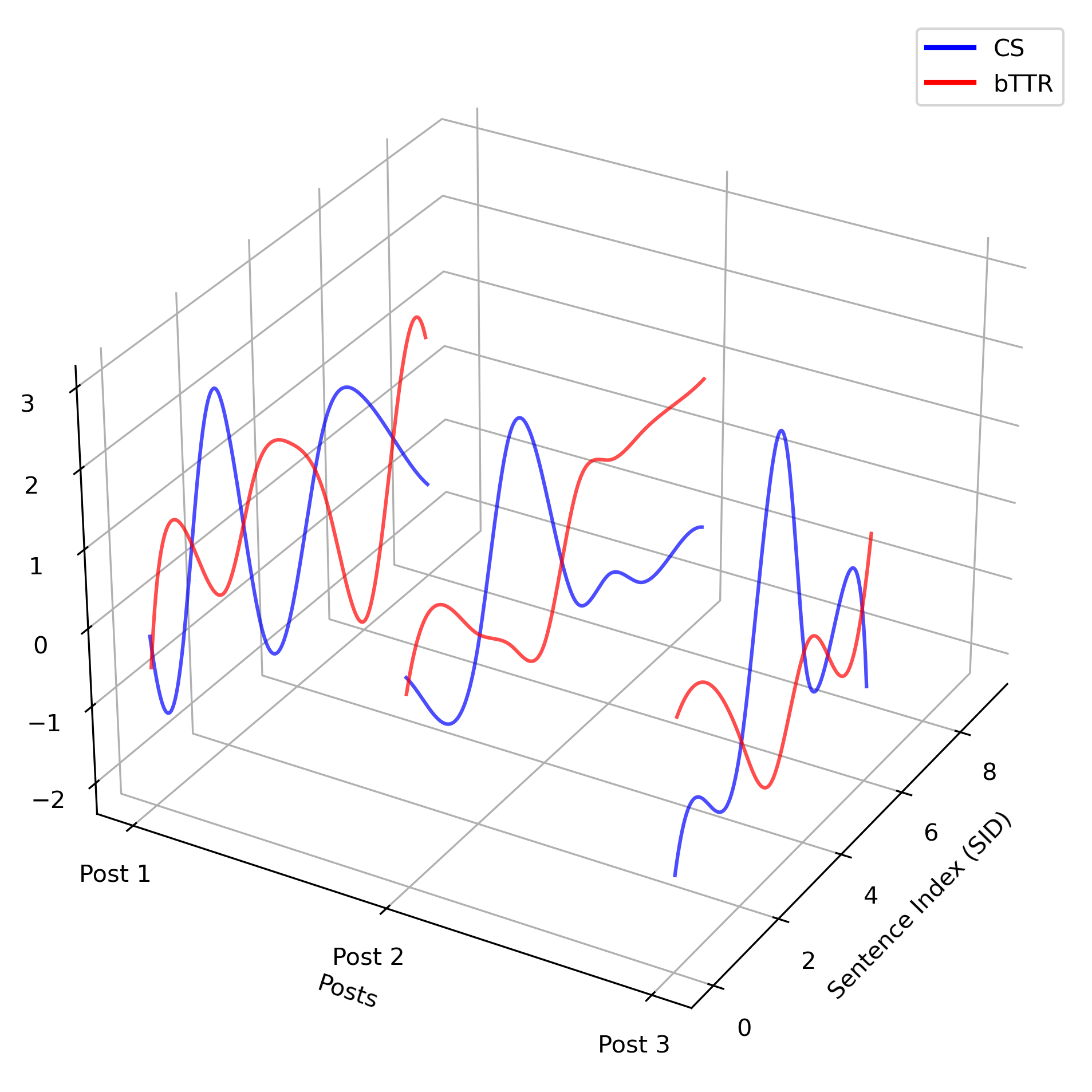}
  \caption {Text Feature Dynamics Across Social Media Posts. The figure illustrates the dynamic contours of text features across a series of measurements, focusing on Clauses per Sentence (CS) and Bilogarithmic Type-Token Ratio (bTTR). It visualizes temporal variations in the usage of these features within three consecutive social media posts from (a) an individual diagnosed with ADHD (left) and (b) a control user (right). The z-standardized scores for each sentence depict the dynamic nature of text usage across the posts.}
  \label{fig:CYMO}
\end{figure*}

CYMO employs the spaCy library for core tasks, including tokenization, sentence segmentation, part-of-speech (POS) tagging, lemmatization, and syntactic parsing. The output from spaCy feeds into CYMO’s measurement module, where feature values are computed for each text. CYMO implements a sliding-window technique to generate high-resolution measurements to capture feature distributions at a granular level, processing the text sentence-by-sentence to ensure detailed, localized insights throughout the text, as illustrated in Figure \ref{fig:CYMO}.

\subsection{ADHD Detection Models}

For the experiments with the \textbf{shallow machine learning models}, we employed four classifiers: Logistic Regression, Random Forest, Support Vector Machine (SVM), and Gradient Boosting. The input to these models consists of the mean feature scores for each user, averaged over all sentences in all their posts. All features were standardized using a StandardScaler prior to training. For Logistic Regression, we used an elastic net penalty with C = 0.02 and an l1-ratio of 0.05, optimized with the SAGA solver. The Random Forest model was configured with 56 trees, a maximum depth of 13, and a minimum sample split of 6, using the square root of the number of features for splitting. The SVM model employed a radial basis function (RBF) kernel with a regularization parameter C = 1.8 and gamma set to scale. For Gradient Boosting, the model was set with 45 estimators, a learning rate of 0.4, and a maximum tree depth of 3.

For the \textbf{deep learning experiments} using CYMO features, we implemented a 3-layer Bidirectional Long Short-Term Memory (BiLSTM) model with a hidden state dimension of 256 units. Each user's input data consisted of the concatenation of all their posts, represented as a sequence \(\{x_1, x_2, \dots, x_N\}\), where each post \(x_i\) is composed of sentences, each characterized by a vector of 344 features:

\[
x_i = \{s_1, s_2, \dots, s_L\}, \quad s_j \in \mathbb{R}^{344}
\]

In the final layer of the BiLSTM, the last hidden states from both the forward \(\overrightarrow{h_3}\) and backward \(\overleftarrow{h_3}\) passes were extracted and concatenated into a vector that represents the entire input sequence:

\[
h_3 = [\overrightarrow{h_3} \parallel \overleftarrow{h_3}]
\]

This concatenated representation was fed into a 3-layer feedforward neural network, where each hidden layer comprised 512 units. A Parametric Rectified Linear Unit (PReLU) activation function was used to introduce non-linearity, and the final output layer employed a sigmoid activation function for binary classification.

The model was trained using binary cross-entropy loss, optimized via backpropagation through time (BPTT). We used the AdamW optimizer, with \(\epsilon = 1 \times 10^{-8}\) and \(\beta\) values of \((0.9, 0.999)\). A OneCycleLR scheduler was applied, with a maximum learning rate of 0.01. The learning rate increased during the first 30\% of the training cycle and then gradually decreased for the remaining 70\%, settling at a lower value determined by a final division factor of \(1 \times 10^{4}\). Training was conducted for 60 epochs, with a dropout rate of 0.2 applied to all 3 recurrent layers to prevent overfitting. Early stopping was employed based on validation performance to enhance generalization. The input sequence length was capped at 200 sentences to efficiently handle longer texts.

In our experiments with transformer-based models, we implemented the architectures using PyTorch \citep{paszke2019pytorch} and the Hugging Face Transformers library \cite{wolf2019huggingface}. Both RoBERTa \cite{Liu2019RoBERTa} and MentalRoBERTa \cite{ji2021mentalbert} use the \texttt{roberta-base} encoder. Each model consists of 12 layers, 12 attention heads, and a hidden representation dimensionality of 768. Input texts were tokenized with their respective tokenizers, with a maximum sequence length of 512 tokens, the upper limit for input sequences in BERT-based architectures.

For each post, a post-level representation \(p\) was derived by extracting the hidden state of the [CLS] token:

\[
p = \text{Model[CLS]}(w_1, w_2, \dots, w_L)
\]

where \(w_1, w_2, \dots, w_L\) are the words in the post. To capture relationships between multiple posts from a user, a user-level encoder based on the transformer architecture \cite{NIPS2017_3f5ee243} was employed. This encoder models the relationships between post representations \(\{p_1, p_2, \dots, p_K\}\), where \(K\) represents the number of posts. The encoder outputs updated post representations \(\{p'_1, p'_2, \dots, p'_K\}\), which are used for downstream tasks such as binary classification.

We applied a uniform training methodology for all transformer-based models to ensure consistency across architectures. A batch size of 16 is used for both training and evaluation. The models are optimized using the AdamW optimizer \cite{loshchilov2017decoupled}, with a learning rate initially set to 0 and increased linearly to 2e-5 over the first 10\% of total training steps, followed by a linear decay schedule, along with a weight decay of 1e-5. The transformer encoders output a 768-dimensional vector, which is processed through a fully connected linear layer and a sigmoid activation function to compute the probability of the positive class (ADHD). Training employs binary cross-entropy loss with logits (BCEWithLogitsLoss), and dropout is applied at a rate of 0.1 to mitigate overfitting. Each model is fine-tuned for 5 epochs, with gradient accumulation, saving the version with the lowest validation loss at each epoch to retain the best-performing model for inference.

\textbf{Hyperparameter tuning} was conducted on the development set, using F1 scores as the evaluation metric. For the shallow machine learning models (Logistic Regression, Random Forest, SVM, and Gradient Boosting) and the MentalRoBERTa model, we utilized grid search to systematically explore a range of hyperparameter combinations. For the BiLSTM models, a sequential tuning approach was applied, progressively adjusting hyperparameters based on validation performance. This method allowed us to iteratively refine the model complexity, ensuring optimal performance across all models.


\section{Results and Analysis}

This section presents a detailed analysis of the performance of several machine learning models, including shallow models (Gradient Boosting, Random Forest, Support Vector Machines, and Logistic Regression), deep learning models (BiLSTM), and transformer-based models (RoBERTa and MentalRoBERTa), in detecting ADHD from social media data. We begin by outlining the results of in-domain classification, where models were trained and evaluated on the same dataset. This is followed by a feature ablation study that highlights the importance of specific NLP-derived features for ADHD detection. Finally, we assess the generalizability of the models using an out-of-domain dataset, examining how well the models perform on unseen data.

\subsection{In-Domain Classification Results}

The results of the in-domain ADHD detection experiments are summarized in Table \ref{tab:idresults}, presenting evaluation metrics of precision, recall, and F1-score. Among the shallow machine learning models, Gradient Boosting (GB) emerged as the best performer, achieving an F1-score of 0.76. The Support Vector Machine (SVM) also attained an F1-score of 0.76 but exhibited a slightly lower recall of 0.75. The Random Forest (RF) model followed closely with an F1-score of 0.75, while Logistic Regression (LR) achieved an F1-score of 0.74.  The Bi-directional Long Short-Term Memory (BiLSTM) model achieved the highest F1-score of 0.77, matching the performance of M-RoBERTa. Additionally, it recorded the highest recall (0.79) among all models. With a precision of 0.75, BiLSTM demonstrated a strong balance between precision and recall in detecting ADHD instances.

The transformer models RoBERTa and MentalRoBERTa demonstrated similar performance, each achieving a Precision of 0.75. RoBERTa recorded a Recall of 0.77 and an F1-score of 0.76, while MentalRoBERTa slightly outperformed it with a Recall of 0.79 and an F1-score of 0.77. This result underscores the value of domain-specific fine-tuning in enhancing model performance, consistent with prior research on binary classification models for mental disorders, including depression, anxiety and suicidal ideation \cite{ji2021mentalbert}. Although the improvement is modest, it suggests that adapting models to domain-specific data enhances their ability to detect subtle patterns, which is crucial in specialized tasks like ADHD classification.

Overall, the findings indicate that the best-performing models for ADHD detection are BiLSTM and MentalRoBERTa, each achieving an F1-score of 0.77. However, the BiLSTM model offers a notable advantage in transparency by enabling feature ablation. This approach uncovers significant NLP-derived insights from verbal behavior that inform model decisions (see Section 4.2) and aids in identifying potential digital biomarkers that surpass the interpretability limitations of MentalRoBERTa.

\begin{table}[ht]
    
    \label{tab:performance_metrics}
    \centering
    \begin{tabularx}{\columnwidth}{@{}Xccc@{}}
        \toprule
        Model & Precision & Recall & F1-score \\
        \midrule
        LR & 0.75 & 0.74 & 0.74 \\
        RF & 0.78 & 0.73 & 0.75 \\
        SVM & 0.77 & 0.75 & 0.76 \\
        GB & 0.78 & 0.75 & 0.76 \\
        BiLSTM & 0.75 & 0.79 & 0.77 \\
        \midrule
        RoBERTa  & 0.75 & 0.77 & 0.76 \\
        M-RoBERTa & 0.75 & 0.79 & 0.77 \\
        \bottomrule
    \end{tabularx}
    \caption{Performance of models on Reddit data, showing precision, recall, and F1-score for in-domain ADHD detection. These metrics are reported for the positive class representing ADHD instances.}
    \label{tab:idresults}
\end{table}

\subsection{Feature Ablation Results}

We conducted feature ablation experiments using Submodular Pick LIME (SP-LIME), an extension of LIME designed to provide global explanations for models. SP-LIME achieves this by selecting a representative subset of local linear approximations through a submodular optimization algorithm \cite{ribeiro2016should}. These global explanations are generated by aggregating the weights of local models that approximate the original model's behavior.

To begin, local explanations were generated using LIME by perturbing the input data and constructing linear models to explain individual predictions. We represented the presence or absence of feature groups in the perturbed samples with binary vectors \( z \in \{0,1\}^d \), where \( d \) is the number of feature groups. An exponential kernel based on Hamming distance, with a kernel width \( \sigma = 0.75 \sqrt{d} \), was applied to weight each perturbed sample differently. The global feature importance score for each feature group \( j \) was then computed as:

\[
I_j = \sqrt{\sum_{i=1}^{n} |W_{ij}|}
\]

where \( n \) is the number of perturbed samples, and \( W_{ij} \) is the coefficient of feature group \( j \) in the linear model fitted to perturbed sample \( i \).

To complement the interpretation of the SP-LIME ablation results, we also employed Mean Decrease in Impurity (MDI) from Random Forest models as a secondary method \cite{breiman2001random}. The outcomes of the SP-LIME feature ablation experiments are summarized in Table \ref{tab:splime_results}.

\begin{table}[]
\centering
\label{tab:splime_results}
\begin{tabularx}{\columnwidth}{Xc}
    \toprule
    Feature Group                & SP-LIME I-value \\
    \midrule
    Readability                  & 27.486 $\downarrow$      \\
    Grammatical Categories       & 18.873 $\uparrow$        \\
    Topical Categories           & 16.847 $\downarrow$      \\
    Stylistics                   & 15.442 $\uparrow$        \\
    Cohesion                     & 13.495 $\uparrow$        \\
    Syntactic Complexity         & 12.609 $\uparrow$        \\
    Lexical Complexity           & 11.915 $\uparrow$        \\
    Emotion Categories           & 8.651 $\uparrow$         \\
    \bottomrule
\end{tabularx}
\caption{SP-LIME feature ablation results showing the I-values for the eight feature groups. The upward arrow (\(\uparrow\)) indicates that the majority of features within the respective group have higher mean scores in the ADHD group compared to the control group, while the downward arrow (\(\downarrow\)) indicates that the majority of features have lower mean scores in the ADHD group.}
\label{tab:splime_results}
\end{table}

The feature ablation results indicate that the most important feature group is \textit{Readability} (I-value: 27.486), followed by \textit{Grammatical Categories} (18.873), \textit{Topical Categories} (16.847), \textit{Stylistics} (15.442), \textit{Cohesion} (13.495), \textit{Syntactic Complexity} (12.609), \textit{Lexical Complexity} (11.915), and \textit{Emotion Categories} (8.651). The downward arrows ($\downarrow$) for \textit{Readability} and \textit{Topical Categories} reflect lower mean scores in the verbal behavior of individuals with ADHD, whereas the upward arrows ($\uparrow$) for \textit{Grammatical Categories}, \textit{Stylistics}, \textit{Cohesion}, \textit{Syntactic Complexity}, \textit{Lexical Complexity}, and \textit{Emotion Categories} indicate higher mean scores. Due to space limitations, we focus our discussion on a few representative examples of these feature groups. 

The reduced readability scores observed in the language production of individuals with ADHD, in conjunction with heightened syntactic and lexical complexity, may reflect the cognitive and behavioral traits characteristic of the condition. For instance, hyperactivity and impulsivity could contribute to more rapid or verbose language production, resulting in structurally complex sentences and broader vocabulary usage. Additionally, cognitive features such as divergent thinking and hyperfocus, which are often associated with ADHD, might promote the production of more elaborate linguistic expressions. Compensatory strategies, such as verbal self-regulation, could also play a role as individuals strive to maintain attention and organize their thoughts through more complex linguistic forms. These observations align with ADHD-related cognitive traits as described in the DSM-5-TR \cite{APA2022}, though further research is needed to explore these mechanisms.

The findings from the MDI analysis indicate that language production among individuals with ADHD is characterized by an increased use of self-referential pronouns within grammatical categories. This observation is consistent with previous research \cite{guntuku2019language}. One possible explanation for this phenomenon is the high comorbidity between ADHD and bipolar disorder \cite{schiweck2021comorbidity}, the latter of which is often associated with rumination—a form of self-focused attention that involves repetitive and persistent negative thinking centered around the self \cite{beck2024cognitive,schiweck2021comorbidity}. These cognitive patterns may influence language use, particularly the heightened self-referencing observed in ADHD-related speech.

Turning to the emotional categories, the MDI analysis revealed an elevated frequency of negative emotions, including the sub-categories of anger, fear, and disgust, which aligns with prior research \cite{guntuku2019language}. This increase in negative emotional content suggests that individuals with ADHD may experience more intense or frequent negative affect, likely related to heightened emotional dysregulation—a central feature of ADHD \cite{shaw2014emotion,thorell2020emotion}. These findings highlight the role of emotional regulation difficulties in ADHD and their potential impact on emotional expression.

Shifting focus to the topical categories, the analysis revealed that individuals with ADHD tend to address more health-related topics in their social media posts compared to neurotypical individuals. This is reflected in the higher proportion of words from health-related categories, which aligns with previous research suggesting that individuals with ADHD frequently discuss personal well-being and health concerns on digital platforms \cite{guntuku2019language, kalantari2023understanding}. These tendencies may be attributed to the challenges that individuals with ADHD face in managing both physical and mental health, as well as a heightened need to seek information or support regarding their symptoms and treatments.

\subsection{Generalizability}

To evaluate the generalizability of our models, we conducted additional experiments using the Twitter-Self-Reported Temporally-Contextual Mental Health Diagnosis Dataset \cite[Twitter-STMHD][]{singh2022twitter}, a large-scale social media dataset collected from Twitter. Twitter-STMHD includes data from 25,860 users who self-reported diagnoses across eight mental health disorders, including ADHD, as well as a control group of approximately 8,000 users without reported mental health conditions. The dataset was constructed using ``anchor tweets,'' where users disclosed their mental health status with phrases such as ``diagnosed with \textless disorder name\textgreater.'' For our out-of-domain (OOD) experiments, we sampled text data from 1,000 users diagnosed with ADHD and an equal number of control users.

\begin{table}[ht]
    
    \label{tab:performance_metrics}
    \centering
    \begin{tabularx}{\columnwidth}{@{}Xcccc@{}}
        \toprule
        Model & P & R & F1 & Change F1\\
        \midrule
        LR          & 0.59 & 0.54 & 0.56 & -0.18 \\
        RF          & 0.63 & 0.76 & 0.69 & -0.06 \\
        SVM         & 0.64 & 0.74 & 0.68 & -0.08 \\
        GB          & 0.62 & 0.82 & 0.70 & -0.06 \\
        BiLSTM      & 0.65 & 0.64 & 0.65 & -0.12 \\
        RoBERTa     & 0.87 & 0.49 & 0.63 & -0.13 \\
        M-RoBERTa & 0.85 & 0.49 & 0.62 & -0.15\\
        \bottomrule
    \end{tabularx}
    \caption{Performance of models on Twitter data, showing the change in F1-score compared to their performance on Reddit data summarized in Table \ref{tab:idresults}.}
    \label{tab:OODresults}
\end{table}

Table \ref{tab:OODresults} summarizes the performance metrics of the models for ADHD detection in the out-of-domain setting, including Precision (P), Recall (R), F1-score, and the Change F1-score compared to their performance on the Reddit dataset (as presented in Table \ref{tab:idresults}). Among the traditional machine learning models, Logistic Regression (LR) achieved an F1-score of 0.56, reflecting a 0.18 drop in performance from Reddit data. Random Forest (RF) and Support Vector Machine (SVM) exhibited better recall, with F1-scores of 0.69 and 0.68, respectively, indicating a drop of 0.06 and 0.08. Gradient Boosting (GB) performed the best among the traditional models, attaining an F1-score of 0.70, with a minimal drop of 0.06. The Bi-directional Long Short-Term Memory (BiLSTM) model demonstrated an F1-score of 0.65, reflecting a 0.12 decrease in performance. The transformer-based models, RoBERTa and MentalRoBERTa (M-RoBERTa), achieved high precision values of 0.87 and 0.85, respectively, but exhibited lower recall rates of 0.49, resulting in F1-scores of 0.63 and 0.62. These results suggest that models trained on what, to the best of our knowledge, represents the most extensive set of human-interpretable expert-engineered features—including Gradient Boosting (GB), Random Forest (RF), Support Vector Machines (SVM), and Bi-directional Long Short-Term Memory (BiLSTM)—exhibit superior robustness in out-of-distribution ADHD classification compared to transformer-based models. With F1-scores ranging from 0.65 to 0.70 and minimal performance degradation, these models demonstrate a commendable balance between accuracy and generalizability.

In contrast, while the transformer-based models, RoBERTa and MentalRoBERTa (M-RoBERTa), exhibit high precision (0.87 and 0.85, respectively), they suffer from significantly lower recall rates (0.49) and thus lower F1-scores (0.63 and 0.62). This trade-off highlights the limitations of transformer architectures in effectively generalizing to out-of-domain data, reinforcing the advantages of employing interpretable, feature-based approaches for real-world applications in ADHD detection.

\section{Conclusion}

 Our work demonstrates that NLP and ML techniques can effectively detect ADHD through social media text, with models like BiLSTM showing strong performance while maintaining interpretability. The results emphasize the importance of balancing accuracy with transparency, particularly for real-world healthcare applications. Feature ablation experiments revealed that linguistic features related to readability, grammatical categories, and emotional expression play a significant role in identifying ADHD-related verbal behavior. Furthermore, out-of-distribution experiments highlight the generalizability challenges faced by transformer-based models, reinforcing the value of interpretable, feature-based approaches for reliable ADHD detection across diverse contexts. These findings pave the way for the development of robust and scalable digital screening tools for ADHD.

\section{Ethical Consideration}

We utilize publicly available Reddit posts for our analysis, implementing stringent privacy measures throughout the data collection process\footnote{According to Reddit's privacy policy, posts are public and accessible to everyone. Reddit allows third parties to access public content via its API. For details, see \url{https://www.reddit.com/policies/privacy-policy}.}. We follow established guidelines \cite{cohan-etal-2018-smhd, singh2022twitter} and refrain from engaging with users or associating their activity with other platforms. Usernames are replaced with randomized identifiers to ensure anonymity. For datasets related to symptom identification, we comply with all data usage agreements to minimize privacy risks. This work is designed to support experienced clinicians in mental health assessments, not to replace clinical judgment.

\section{Limitations}

Our work has several limitations that could be addressed in future research: (1) \textbf{Focus on a Single Disorder:} This study concentrates on ADHD. Future research should include associated disorders, such as autism spectrum disorder and anxiety disorder, to facilitate differential analysis and address comorbidity challenges. (2) \textbf{Generalizability of Findings:} The current analysis relies on social media data. Future studies should incorporate data from clinical settings, such as interviews, to understand differences and similarities in ADHD detection. (3) \textbf{Advancements in LLM-Based Approaches:} The rapid development of large language models (LLMs) offers opportunities for further exploration. Future research could evaluate their performance in detecting ADHD compared to traditional methods.

\bibliography{paper}

\appendix

\newpage
\section{Appendix}
\label{sec:appendix}

\section*{Pre-processing}
All social media posts were preprocessed by converting text to lowercase, removing hyperlinks, HTML tags, and user mentions. Hashtags were decomposed into constituent words, and all non-alphanumeric characters, except for essential punctuation to maintain syntactic structure, were removed. Text normalization corrected spacing and punctuation irregularities, and posts were segmented into individual sentences. Posts with fewer than three sentences were discarded to enable the measurement of textual coherence based on the full feature set supported by CYMO. Deduplication was performed to ensure that only unique and relevant entries remained for further analysis

We use stratified sampling to randomly split the ADHD and control datasets into training, validation, and test sets with a split ratio of 8:1:1, ensuring that the label distribution remains consistent across all sets.

MDI calculates the average decrease in impurity—measured using Gini impurity—resulting from node splits across the ensemble of decision trees. The importance score for each feature \( j \) is computed as follows:

\begin{equation}
I_j = \frac{1}{T} \sum_{t=1}^{T} \Delta \text{Impurity}_t^j
\end{equation}

where \( \Delta \text{Impurity}_t^j \) represents the reduction in impurity attributed to feature \( j \) across tree \( t \) in the ensemble, and \( T \) is the total number of trees.

Table 5 presents the top 25 features evaluated in this work. A table listing the results of all 344 features is provided in the appendix. A substantial majority of the top-25 features belong to the Emotion Categories (44\%) and Cohesion (24\%), followed by Stylistics (16\%), Grammatical Categories (12\%), and Topical Categories (4\%). Individuals diagnosed with ADHD tend to utilize a higher frequency of terms associated with negative emotions, such as loathing, disgust, terror, rage, grief, anger, and fear. Additionally, they demonstrate increased lexical overlap across adjacent sentences, particularly in their use of function words. This group also shows a greater proportion of self-referential pronouns and frequently employs n-grams characteristic of both fiction and internet discourse, along with an elevated use of vocabulary related to health topics.

\begin{table*}[htbp]
\centering
\caption{MDI for top 25 features evaluated in this work}
\begin{tabular}{@{}lllc@{}}
\toprule
Feature   & Name                                       & Category                 & Importance   \\ \midrule
EMOloa    & Emotion Loathing                           & Emotion Categories       & 0.024843427  \\
N2SFWO    & Next-Two Sentences Function Word Overlap   & Cohesion                 & 0.018468562  \\
EMOdsg    & Emotion Disgust                            & Emotion Categories       & 0.014262222  \\
N2SPO     & Next-Two Sentences Pronoun Overlap         & Cohesion                 & 0.014200314  \\
EMOter    & Emotion Terror                             & Emotion Categories       & 0.013693997  \\
EMOrag    & Emotion Rage                               & Emotion Categories       & 0.01296565   \\
EMOgri    & Emotion Grief                              & Emotion Categories       & 0.011748314  \\
PRNref1s  & Pronoun (reflexive, 1st, sg)               & Grammatical Categories   & 0.011429512  \\
TOPhea    & Topic Health                               & Topical Categories       & 0.011312342  \\
N2SLO     & Next-Two Sentences Lemma Overlap           & Cohesion                 & 0.011260935  \\
EMOanx    & Emotion Anxiety                            & Emotion Categories       & 0.010800668  \\
NSFWO     & Next-Sentence Function Word Overlap        & Cohesion                 & 0.010560373  \\
EMOann    & Emotion Annoyance                          & Emotion Categories       & 0.010274373  \\
EMOang    & Emotion Anger                              & Emotion Categories       & 0.009983665  \\
EMOfea    & Emotion Fear                               & Emotion Categories       & 0.009691779  \\
PRNref    & Pronoun (reflexive)                        & Grammatical Categories   & 0.009563418  \\
EMOdsl    & Emotion Dislike                            & Emotion Categories       & 0.008679112  \\
3GNLFf    & Trigram Fiction Normalized Log Frequency   & Stylistics               & 0.008137377  \\
2GNLFf    & Bigram Fiction Normalized Log Frequency    & Stylistics               & 0.007098874  \\
NSAdvO    & Next-Sentence Adverb Overlap               & Cohesion                 & 0.006957162  \\
2GNLFb    & Bigram Weblog Normalized Log Frequency     & Stylistics               & 0.006804482  \\
NSPO      & Next-Sentence Pronoun Overlap              & Cohesion                 & 0.006788781  \\
EMOmel    & Emotion Melancholy                         & Emotion Categories       & 0.006535816  \\
3GNLFw    & Trigram Web Normalized Log Frequency       & Stylistics               & 0.005564999  \\
DETposs1p & Determiner (possessive, 1st, pl)           & Grammatical Categories   & 0.005455105  \\ 
\bottomrule
\end{tabular}
\end{table*}

\end{document}